\pdfoutput=1
\documentclass[11pt]{article}

\usepackage{acl}

\usepackage{times}
\usepackage{graphicx}
\usepackage{latexsym}
\usepackage{CJKutf8}
\usepackage{graphicx}
\usepackage{colortbl}
\usepackage{tikz}

\usepackage{colortbl}
\usepackage{verbatim}
\usepackage{amsmath} 
\usepackage[normalem]{ulem}
\usepackage{multirow}
\usepackage{rotating}
\usepackage{caption}
\usepackage{subcaption}
\usepackage{tabularx}
\usepackage{booktabs}
\usepackage{supertabular}
\usepackage{subcaption}
\usepackage{arydshln}

\usepackage[utf8]{inputenc}
\usepackage[T1]{fontenc}
\usepackage{microtype}
\usepackage{inconsolata}

\title{Empowering Multi-step Reasoning across Languages via Tree-of-Thoughts}

\author{Leonardo Ranaldi$^{(\diamond)}$, Giulia Pucci $^{(\star)}$, \\
\textbf{Federico Ranaldi $^{(\diamond)}$, Elena Sofia Ruzzetti $^{(\diamond)}$, Fabio Massimo Zanzotto$^{(\diamond)}$} \\
 ${(\diamond)}$Human-Centric ART Group,\\ Dep. of Enterprise Engineering, University of Rome Tor Vergata, Italy \\
  ${(\star)}$Department of Computing Science, University of Aberdeen, UK\\
    \tt [first\_name].[last\_name]@uniroma2.it	
}

\begin{document}
\maketitle
\begin{abstract}

Reasoning methods, best exemplified by the well-known Chain-of-Thought (CoT),  empower the reasoning abilities of Large Language Models (LLMs) by eliciting them to solve complex tasks in a step-by-step manner. Although they are achieving significant success, the ability to deliver multi-step reasoning remains limited to English because of the imbalance in the distribution of pre-training data, which makes other languages a barrier. \\
In this paper, we propose Cross-lingual Tree-of-Thoughts (Cross-ToT), a method for aligning Cross-lingual CoT reasoning across languages. The proposed method, through a self-consistent cross-lingual prompting mechanism inspired by the Tree-of-Thoughts approach, provides multi-step reasoning paths in different languages that, during the steps, lead to the final solution. Experimental evaluations show that our method significantly outperforms existing prompting methods by reducing the number of interactions and achieving state-of-the-art performance.
\end{abstract}

\section{Introduction}
Chain-of-Thought (CoT) prompting elicits Large Language Models (LLMs) to break down a reasoning task towards a sequence of intermediate steps \cite{wei2022emergent}. 
Previous works have demonstrated that LLMs achieve impressive performances in zero-shot learning scenarios without the need to modify the model parameters during the training and testing process.
In particular, by appending to the prompt “Let’s think step by step!” \cite{kojima2023large} LLMs with at least several billions of parameters, such as GPTs family \cite{openai2023gpt4} or PaLM \cite{chowdhery2022palm}, deliver multi-step controlled reasoning, achieving promising results across commonsense \cite{bubeck2023sparks}, symbolic and mathematical reasoning datasets \cite{gaur-saunshi-2023-reasoning,liu2023evaluating}.

Although the performances seem promising, they are only firmly established in English. This poses a barrier to generalizing current CoT techniques to different languages.
Hence, despite the remarkable success of zero-shot CoT techniques, the reasoning abilities of LLMs still struggle to generalize to different languages. \citet{shi2022language} introduced the first multilingual benchmark to assess LLMs' mathematical reasoning abilities using prompts in different languages. \citet{qin2023crosslingual} propose task-specific solver prompting, using a succession of prompts, elicit the LLMs to understand questions and deliver CoT answers in different languages. However, these strategies require two-step prompts, which goes against the zero-shot approach.

In this paper, we propose Cross-lingual Tree-of-Thoughts (Cross-ToT), a method for aligning Cross-lingual CoT reasoning across languages by proposing a Cross-lingual Alignment prompt to elicit the model to deliver a Self-consistent Chain-of-Thougt.
Our method is inspired by the Tree-of-Thoughts (ToT) prompting \cite{yao2023tree} that asks LLMs to perform decision-making by considering multiple different reasoning paths (CoTs).
In particular, our Cross-ToT is a ToT-style prompting to deliver the reasoning process in different languages that, step-by-step, converge to a single final solution. The inherent insight is that as the different paths of thought evolve, the relationships between the different languages are inherently grasped via Self-consistent Chains-of-Thougt. This leads to the target research questions, which are the focus of this paper: 

\textit{RQ1:} Are LLMs able to deliver Cross-lingual multi-step reasoned answers?

\textit{RQ2:} Are the different paths of ToT evolving Self-correcting each other?

\textit{RQ3:} What is the role of English in Cross-lingual scenarios?

\begin{figure*}[t!]
\centering
    \includegraphics[width=1.0\textwidth]{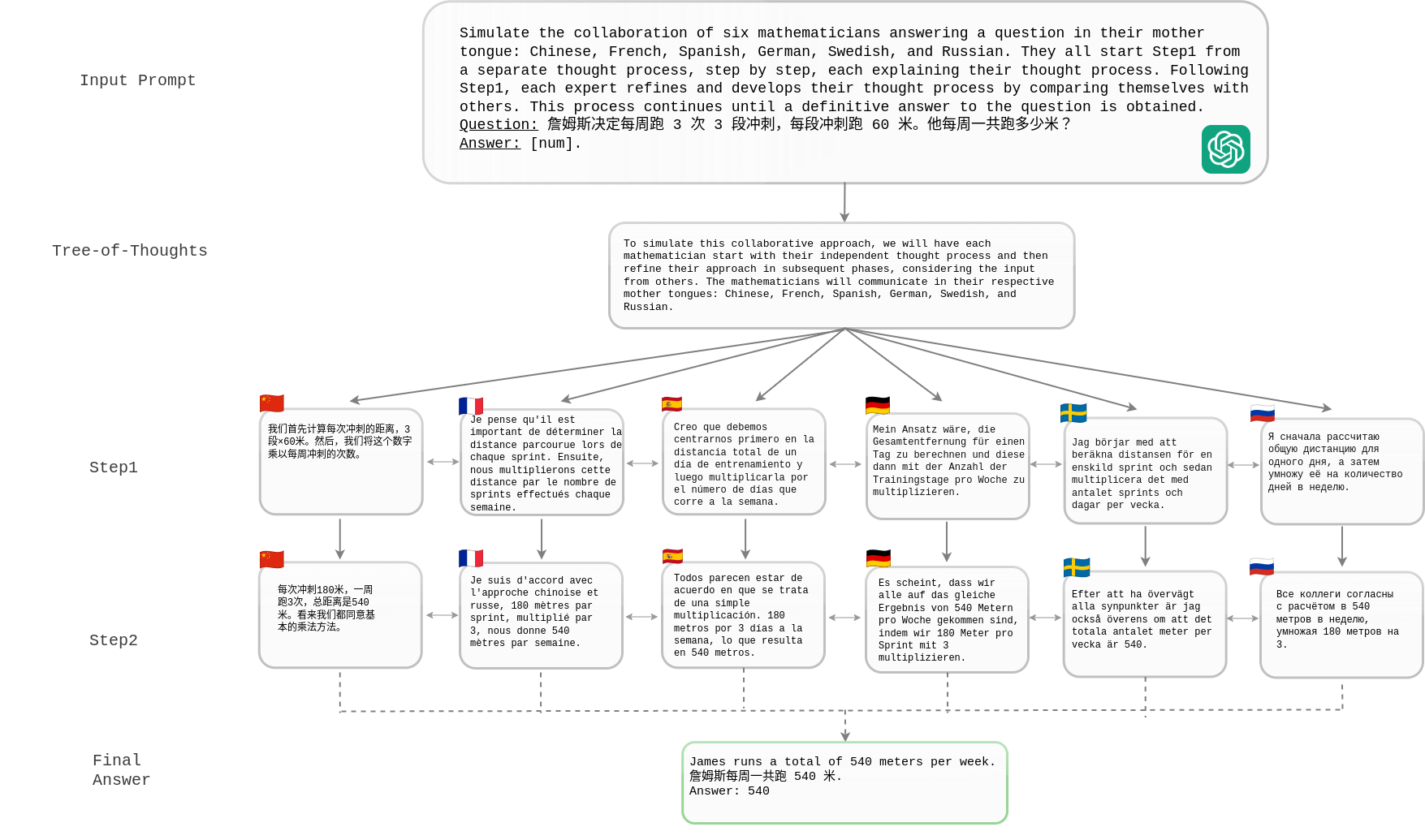}
    \caption{Our \texttt{Cross-ToT} elicits the LLM to generate step-by-step Cross-lingual reasoning. Furthermore, different pathways are developed during these reasoning steps. This mechanism develops the Chains-of-Thoughts in a Self-consistent way, streaming with the different pathways.}
    \label{fig:our_proposal}
\end{figure*}

To answer these questions, we propose \texttt{Cross-ToT}, a novel Cross-lingual prompting strategy that aims to bridge the gap across different languages. In particular, using the prompt shown in Figure \ref{fig:our_proposal}, we elicit the model to deliver different CoT reasoning steps in different languages that converge to the final solution step-by-step. We test our method on GPT-3.5 and conduct an extensive analysis using Multilingual Grade School Math (MGSM) \cite{shi2022language}, Cross-lingual Natural Language Inference (XNLI) \cite{conneau-etal-2018-xnli}, and Cross-lingual Paraphrase Adversaries Scrambling (PAWS-X) \cite{yang-etal-2019-paws}, Cross-lingual Choice of Plausible Alternatives (XCOPA) \cite{ponti-etal-2020-xcopa} across different languages. 
Experimental results reveal that our method, based on a single prompt, outperforms the baselines and achieves the SOTA performance on different languages in different tasks. 
The main contributions of this work are concluded as follows:

\begin{itemize}
\item We introduce \texttt{Cross-ToT}, which is a novel Cross-lingual prompting mechanism that stimulates the model to produce parallel CoT reasoning processes across different languages;
\item We show that our \texttt{Cross-ToT} is Self-consistent and allows the integration of reasoning paths between different languages;
\item Extensive evaluations on different languages demonstrate that our \texttt{Cross-ToT} can effectively improve the performance of cross-lingual CoTs and achieve SOTA performance.
\item Finally, we show that introducing English in our prompting technique plays a beneficial role in improving downstream performance. 
\end{itemize}

\section{Cross-lingual Multi-step Reasoning}

To elicit the multi-step reasoning abilities of LLMs in Cross-lingual scenarios, we propose \texttt{Cross-ToT}, which is a Cross-lingual Alignment Chain-of-Thought as a solution. In particular, our method overcomes the Multi-lingual and Cross-lingual approaches introduced in Section \ref{sec:CrossAlignment}. In fact, our approach elicits the LLMs to deliver Self-consistent Parallel Chain-of-Thougts, introduced in Section~\ref{sec:Self-CoT}.

\begin{table}[h]
\small
\centering
\begin{tabular}{|p{0.9\linewidth}|}
\multicolumn{1}{l}{\texttt{\textbf{Native-CoT}} in this example in Chinese} \\
\hline
\begin{CJK*}{UTF8}{gbsn}
\textbf{问题:} 利亚有 32 块巧克力，她妹妹有 42 块。如果她们吃了 35 块，她们一共还剩下多少块？
\end{CJK*} \\
\begin{CJK*}{UTF8}{gbsn}
答案: 让我们一步步思考
\end{CJK*} 
\\
\hline
\end{tabular}
\vspace{0.3em} 
\begin{tabular}{|p{0.9\linewidth}|}
\multicolumn{1}{l}{\texttt{\textbf{En-CoT}}} \\
\hline
\begin{CJK*}{UTF8}{gbsn}
\textbf{问题:} 利亚有 32 块巧克力，她妹妹有 42 块。如果她们吃了 35 块，她们一共还剩下多少块？
\end{CJK*} \\
\texttt{Answer: Let's think step by step} \\ 
\hline
\end{tabular}
\vspace{0.3em} 
\begin{tabular}{|p{0.9\linewidth}|}
\multicolumn{1}{l}{\texttt{\textbf{Translated-CoT}} (is the \texttt{Native} translated in En)} \\
\hline
\texttt{Question: Leah has 32 chocolates and her sister has 42. If they ate 35 pieces, how many pieces do they have left?}\\
\texttt{Answer: Let's think step by step} \\ 
\hline
\end{tabular}
\caption{Different types of input prompts in order to elicit Chain-of-Thought reasoning process. Specifically, given a problem in Chinese, the following prompts are \texttt{Native-CoT} and \texttt{En-CoT}, the original question in Chinese with elicitation in Chinese and English; for \texttt{Translated-CoT}, the question is in English and consequently a step-by-step solution in English.}
\label{tab:example_input_CoT}
\end{table}

\subsection{Chain-of-Thought Across Languages}
\label{sec:CrossAlignment}
The Cross-lingual Alignment is a core challenge for
cross-lingual transfer. \citet{shi2022language} proposed a series of prompts to elicit models to generate CoT answers in specific language \texttt{Native-CoT}, and in English \texttt{En-CoT} and \texttt{Translate-CoT} (more detailed in Table \ref{tab:example_input_CoT}).

Later, \citet{qin2023crosslingual} proposed a method based on two phases: Cross-lingual alignment prompt and task-specific solver prompting. This approach uses two separate steps, as shown in Table \ref{tab:example_input_Cross_CoT}, in order to handle input and output in different languages.

\begin{table}[h]
\small
\centering
\begin{tabular}{|p{0.9\linewidth}|}
\multicolumn{1}{l}{\texttt{\textbf{Cross-CoT} First-Step}} \\
\hline
\texttt{Please act as an expert in multi-lingual understanding in [Specific Language $L_s$]. }\\
\texttt{Question: [Given sentence $X$ in $L_s$]}\\
\texttt{Let’s understand the task in [Target Language $L_t$] step-by-step!} \\ 
\hline
\end{tabular}
\vspace{0.5em}
\begin{tabular}{|p{0.9\linewidth}|}
\multicolumn{1}{l}{\texttt{\textbf{Cross-CoT} Second-Step}} \\
\hline
\texttt{After understanding, you should act as an expert in mathematics in [Language $L_t$].}\\
\texttt{Let’s resolve the task you understand above step-by-step!} \\ 
\hline
\end{tabular}
\caption{Cross-lingual Prompt proposed in \cite{qin2023crosslingual}. By setting an input language and a target language, the prompt is divided into two phases: in phase one, there is the alignment of the different languages, and in phase two, there is the solving mechanism for the specific language.}
\label{tab:example_input_Cross_CoT}
\end{table}

Although this second approach overcomes the limitations of \citet{shi2022language}'s work, the two-step prompting could be more laborious and challenging, and there is no exchange of information during the multi-step reasoning process between the different chains as the final outputs are estimated using a voting heuristic.

\subsection{Self-consistent Parallel Chain-of-Thougts}
\label{sec:Self-CoT}

In our work, we propose \texttt{Cross-ToT}, a prompting method that can handle different languages in a parallel way. 
Furthermore, through a mechanism inspired by Tree-of-Thoughts prompting techniques \cite{yao2023tree}, our method elicits the LLM to deliver the generation of the answer in a sequence of intermediate steps that do not provide independent parallel answers but deliver collaborative Self-consistent reasoned steps until arriving at a final answer.

\begin{table}[h]
\small
\centering
\begin{tabular}{|p{0.9\linewidth}|}
\multicolumn{1}{l}{\texttt{\textbf{Our Proposal}}} \\
\hline
\texttt{Simulate the collaboration of $\{n\}$ mathematicians answering a question in their mother tongue: $L_1$, $L_2$, ... and $L_n$. They all start Step1 from a separate thought process, step by step, each explaining their thought process. Following Step1, each expert refines and develops their thought process by comparing themselves with others. This process continues until a definitive answer to the question is obtained. } \\
\texttt{Question: [Question in Language $L_1$]} \\
\texttt{Answer: [num].} \\ 
\hline
\end{tabular}
\caption{Input-prompt for MSGM task. In \texttt{Cross-ToT}, we elicit the model to produce multi-step reasoning processes in different languages. We specifically prompt to start from separate reasoning and collaborate step-by-step. (We propose similar pattern for other tasks as described in Appendix \ref{sec:prompt})}
\label{tab:example_input_Cross_ToT}
\end{table}

Our \texttt{Cross-ToT} shown in Table \ref{tab:example_input_Cross_ToT} elicits the LLM to generate different paths as shown in Figure \ref{fig:our_proposal}, achieving significant improvements in accuracy as discussed in Section \ref{sec:Results}.

\begin{table*}[t]
\centering
\begin{tabular}{l|cccccccccc|cc}
\textbf{Model} & \textbf{de} & \textbf{zh} & \textbf{fr} & \textbf{ru} & \textbf{sw} & \textbf{es} & \textbf{bn} & \textbf{ja} & \textbf{te} & \textbf{th} & \textbf{Avg} \\ \hline
\textbf{GPT-3 (text-davinci-002)*} &  &  &  &  &  & & & & & &  \\
\texttt{Direct} \cite{shi2022language} & 14.8 & 18.0 & 16.8 & 12.4 & 8.8 & 17.2 & 4.4 & 11.2 & 0.8 & 8.8 & 11.3 \\
\texttt{Native-CoT} \cite{shi2022language} & 36.0 & 40.0 & 37.6 & 28.4 & 11.2 & 40.4 & 6.4 & 26.0 & 0.4 & 10.8 & 23.7 \\
\texttt{En-CoT} \cite{shi2022language} & 44.0 & 40.8 & 46.0 & 28.4 & 20.8 & 44.8 & 9.6 & 32.4 & 5.6 & 19.6 & 29.2 \\
\texttt{Translate-En} \cite{shi2022language} & 46.4 & 47.2 & 46.4 & 48.8 & 37.6 & 51.6 & 41.2 & 44.8  & 42.8 & 41.2 & 44.8 \\
\hdashline
\textbf{GPT-3.5 (gpt-3.5-turbo)} &  &  &  &  &  & & & & & &  \\
\texttt{Direct} \cite{qin2023crosslingual} & 56.0 & 60.0 & 62.0 & 62.0 & 48.0 & 61.2 & 33.6 & 52.8 & 7.6 & 42.2 & 48.5 \\
\texttt{Native-CoT} \cite{qin2023crosslingual} & 70.0 & 59.6 & 64.4 & 62.4 & 54.0 & 70.4 & 26.4 & 64.4 & 40.0 & 59.6 & 57.1 \\
\texttt{En-CoT} \cite{qin2023crosslingual} & 73.6 & 63.2 & 70.0 & 65.6 & 55.2 & 69.6 & 50.0 & 60.4 & 22.0 & 48.0 & 57.7 \\
\texttt{Translate-En} \cite{qin2023crosslingual} & 75.6 & 71.6 & 72.4 & 72.8 & 69.6 & 74.4 & 66.4 & 66.0 & 58.0 & 57.6 & 68.4 \\
\texttt{Cross-CoT} \cite{qin2023crosslingual} & 86.8 & 77.2 & 82.0 & \textbf{87.6} & \textbf{76.0} & 84.8 & 75.2 & 77.2 & 52.0 & 68.0 & 76.6 \\
\hdashline
\textbf{\texttt{Cross-ToT}} & \textbf{87.6} & \textbf{83.5} & \textbf{84.3} & 86.5 & 75.4 & \textbf{86.2} & \textbf{79.0 }& \textbf{80.2} & \textbf{68.5} & \textbf{75.5} & \textbf{80.6} \\
\hline
\end{tabular}
\caption{Accuracies (\%) on MGSM using the \texttt{"Direct"} prompt, i.e., question and answer in the original language; the \texttt{"Native-CoT"} prompt, i.e., question and answer CoT in the original language; the \texttt{"En-CoT"} prompt specific language question and answer CoT in English, the \texttt{"Translate-En"} prompt where the specific input is translated into English and the answer accordingly is in English. Moreover, Cross-CoT, as proposed by \citet{qin2023crosslingual}, questions in a specific language and answers in different languages. Finally, \textbf{\texttt{Cross-ToT}} is explained in Section \ref{sec:Self-CoT}. (Our results are derived from the average of three running performances as detailed in Section \ref{sec:exp_set})}
\label{tab:Results}
\end{table*}

\section{Experiments}
\subsection{Data}
\label{sec:data}
In order to observe the Cross-lingual abilities of LLMs, we used GSM8K \cite{cobbe2021training}, XNLI \cite{conneau-etal-2018-xnli}, and PAWS-X \cite{yang-etal-2019-paws}, XCOPA \cite{ponti-etal-2020-xcopa}. 

\paragraph{Understanding tasks}
In order to assess Cross-lingual comprehension abilities, we used XNLI \cite{conneau-etal-2018-xnli} and PAWS-X. 
The first is an extension of Stanford Natural Language Inference (SNLI) \cite{bowman-etal-2015-large} across 15 languages and is based on one premise and one hypothesis and requires the model to determine whether the hypothesis is entailed, contradicted, or neutral conditioned on the premise in 15 different languages, and we utilize the accuracy score for evaluation.
The second, Paraphrase Adversaries from Word Scrambling (PAWS-X) \cite{yang-etal-2019-paws}, contains two sentences and requires the model to judge whether they paraphrase each other in seven languages.

\paragraph{Commonsense Reasoning task}
The Cross-lingual Choice of Plausible Alternatives (XCOPA) \cite{ponti-etal-2020-xcopa} is based on one premise and two choices. It asks the model to choose which one is the result or cause of the premise. It covers 11 languages from 11 diverse families.

\paragraph{Arithmetic Reasoning task}
To evaluate the problem-solving abilities in Cross-lingual scenarios, we used the extension proposed by \citet{shi2022language}, i.e., Multilingual Grade School Math (MGSM). Initially, \citet{cobbe2021training} proposed a benchmark of mathematical problems in English in GSM8K. Each example has the following structure: a mathematical problem in natural language and a target answer in Arabic number.
\citet{shi2022language}, in their contribution, i.e., MGSM, selected the first 250 examples from the official list of examples in GSM8K and translated them manually into 11 different languages, maintaining the structure of the input and output.

\paragraph{Evaluated Languages}
In our experiments, we propose an analysis of available languages that differ depending on the resources, we provide all details in Appendix \ref{sec:prompt}. 
Furthermore, as an additional experiment, we test the introduction of English.

\subsection{Experimental Setup}
\label{sec:exp_set}
In order to conduct our study on robust models and have a term of comparison with the work proposed in \cite{shi2022language,qin2023crosslingual}, we use GPT-3.5; however, in future developments, we plan to scale the method to different models. 
Then, we systematically defined the input prompt in Table \ref{tab:example_input_Cross_ToT} for MGSM and in Appendix \ref{sec:prompt} for XNLI, PAWS-X, and XCOPA. 
In each particular experimental set-up, we modify the appropriate languages with $L_1$, $L_2$, ...
for the German \footnote{Although we do not observe perceptible changes in the order of languages present in the input prompt, we set as a first the language-related subset of the benchmark.}

Following \citet{wei2022emergent,kojima2023large}, we evaluate performance using the accuracy score. In particular, we compute the string matching between the final answers (see Figure \ref{fig:our_proposal} where the final outputs have the form of \texttt{Answer:[num]}) and the target values. The top-p parameter is set to 1 in all processes. We select the Prompting temperature [0, 1].

\section{Main Results}
\label{sec:Results}

\begin{figure}[t]
\centering
         \begin{minipage}{0.6\linewidth}
     \centering
     \includegraphics[width=\linewidth]{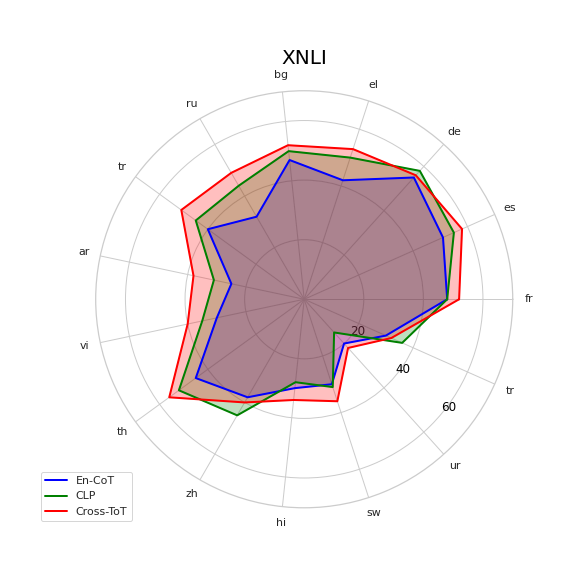}
   \end{minipage}

            \begin{minipage}{0.6\linewidth}
     \centering
     \includegraphics[width=\linewidth]{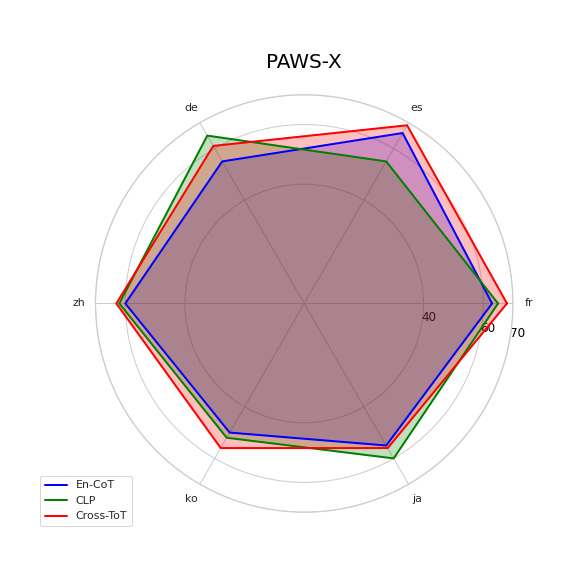}
   \end{minipage}
   
   \caption{Accuracies (\%) on Language Understanding benchmarks XNLI and PAWS-X introduced in Section \ref{sec:data} } 
   \label{fig:performances_Understanding}
\end{figure}

Mechanisms for delivering multistep-reasoned answers across languages can be empowered via \texttt{Cross-ToT} that align languages' Chain-of-Thoughts (CoT).
Our approach based on a Tree-of-Thoughts-inspired prompting mechanism (see Figure \ref{fig:our_proposal}) outperforms state-of-the-art prompting techniques on Arithmetic Reasoning tasks as shown in Table \ref{tab:Results}, and in Language Understanding tasks as shown in Figure \ref{fig:performances_Understanding} and finally in Commonsense Reasoning tasks as shown in Table \ref{tab:XCOPA}.
In particular, \texttt{Cross-ToT} elicit LLMs to produce different reasoning pathways that share the "Thoughts" during the steps and, at the same time, promote Self-correction of mistaken paths.
In fact, during the steps of the CoT, information is swapped between the paths. This interaction delivers Self-consistent paths. Furthermore, in the prompt, we exemplified that the different paths must arrive at a shared and, consequently, unique by sharing the "thought process" (see the prompt in Table \ref{tab:example_input_Cross_ToT}).

\begin{figure}[h]
\centering
    \includegraphics[width=0.50\textwidth]{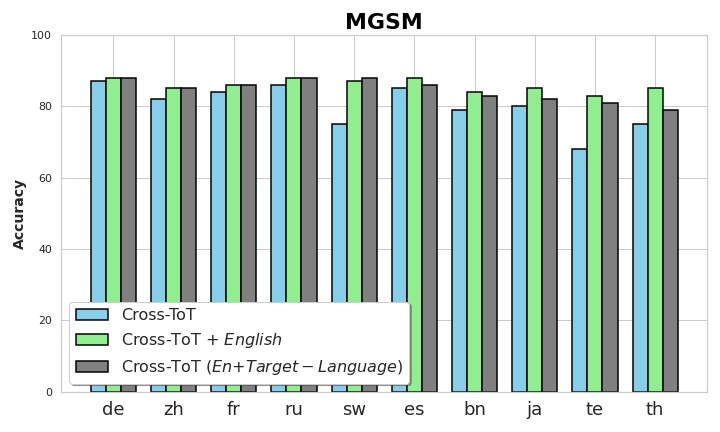}
    \caption{Accuracies (\%) on MGSM using "\texttt{Cross-ToT}", "\texttt{Cross-ToT} + English" and in binary version "\texttt{Cross-ToT} ( English + Target Language".}
    \label{fig:En_ToT}
\end{figure}

Our approach outperforms the methods proposed in \cite{shi2022language} that are yet surpassed by the \texttt{Cross-CoT} proposed by \citet{qin2023crosslingual}. 
However, although \texttt{Cross-CoT} outperforms previous approaches, it is necessary to clarify which path, if any, leads to the correct reasoning (Section \ref{sec:reasoning_evolution}), whether the introduction of English can increase performance (Section \ref{sec:en_matter}) and finally the trade-off between the number of languages (in our case path) and the final results (Section \ref{sec:number_of_languages}).

\section{Analysis}
\label{sec:analysis}
In this section, we explore the contribution of English in the Cross-lingual prompt (in Section \ref{sec:en_matter}), then study the impact of different languages on the final results (Section \ref{sec:number_of_languages}) and the reasoning evolution (Section \ref{sec:reasoning_evolution}) and close with an in-depth analysis of performance in different tasks in Section \ref{sec:other_task}.

\subsection{The English Matter}
\label{sec:en_matter}
Earlier works \cite{wei2022emergent,liu2023evaluating} have been showing that LLMs are able to deliver multi-step reasoning answers on arithmetic tasks, focusing mainly on English.
Therefore, we observe whether introducing English into our input-prompts could increase downstream performance.
Hence, we performed the setting proposed in Section \ref{sec:exp_set}.
From the results obtained in Figure \ref{fig:En_ToT} (green bar), it is possible to observe that the input-prompts empowered with English outperform the input-prompts empowered without English. 
This result suggests that the presence of one robust path, in this case, the English path, may influence the others in the final reasoning process. Indeed, assuming that the production of the intermediate steps is self-consistent, i.e., the paths do not disagree with each other, the additional language seems to influence performance positively. 
Based on the current results, adding further language improves the robustness of the models.

However, whether the performance is due to the number of languages or English is unclear. To observe the impact of adding a specific language in Section \ref{sec:number_of_languages}, we propose to reduce the number of languages in the presence and absence of English.

\begin{table*}[t]
\centering
\small
\begin{tabular}{l|ccccccccccc|cc}
\textbf{Model} & \textbf{et} & \textbf{ht} & \textbf{id} & \textbf{it} & \textbf{qu} & \textbf{sw} & \textbf{ta} & \textbf{th} & \textbf{tr} & \textbf{vi} & \textbf{zh} & \textbf{Avg} \\ \hline

\hline
\textbf{GPT-3 (text-davinci-002)*} & & & & & & & & & &  & &  \\
\texttt{Direct} \cite{shi2022language} & 73.8 & 55.6 & 88.8 & 95.4 & 51.2 & 56.0 & 54.6 & 70.2 & 88.6 & 80.4 & 91.4 & 73.3 \\
\texttt{En-CoT} \cite{shi2022language} & 88.8 & 79.6 & 91.4 & 96.6 & 52.2 & 67.4 & 55.8 & 84.2 & 91.2 & 86.6 & 93.4 & 80.7 \\
\hdashline
\textbf{GPT-3.5 (gpt-3.5-turbo)} & & & & & & & & & &  & &  \\
\texttt{Direct} \cite{qin2023crosslingual}  & 90.6 & 72.0 & 90.4 & 95.2 & 54.6 & 82.0 & 59.0 & 77.6 & 91.0 & 83.6 & 90.4 & 80.6 \\
\texttt{Translate-En} \cite{qin2023crosslingual} & 88.2 & 79.4 & 90.8 & 94.4 & 50.0 & 77.6 & 87.0 & 82.2 & 87.8 & 88.4 & 92.2 & 83.5 \\
\texttt{Cross-CoT} \cite{qin2023crosslingual} & 96.8 & 90.6 & 95.2 & 95.8 & \textbf{85.8} & 92.8 & \textbf{83.2} & 93.2 & \textbf{96.8} & 94.2 & 95.8 & 92.7 \\
\hdashline
\textbf{\texttt{Cross-ToT}} & \textbf{97.6} & \textbf{92.5} & 90.3 & \textbf{96.8} & 83.3 & \textbf{93.6} & 80.2 & \textbf{94.1} & 96.4 & \textbf{95.3} & \textbf{97.4} &  \\
\bottomrule
HUMAN \cite{ponti-etal-2020-xcopa} & 98.2 & 96.4 & 100.0 & 97.0 & 94.8 & 99.0 & 98.6 & 98.2 & 96.4 & 98.4 & 96.6 & 97.6 \\
\end{tabular}
\caption{Accuracies (\%) of XCOPA.}
\label{tab:XCOPA}
\end{table*}

\subsection{The Impact of the Languages}
\label{sec:number_of_languages}
\begin{figure}[h]
\centering
\includegraphics[width=0.55\textwidth]{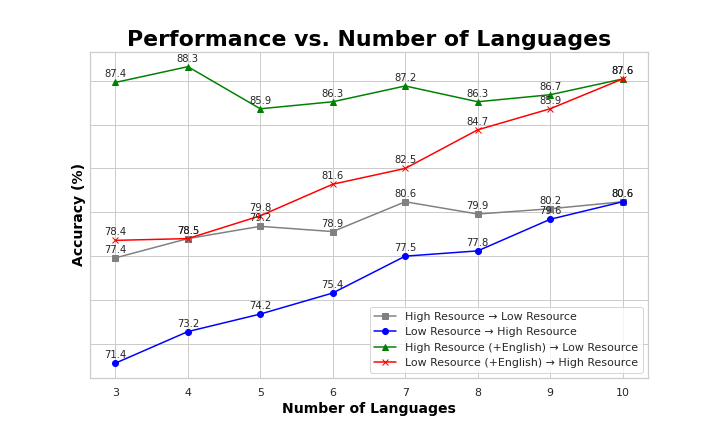}
\caption{The impact of integrating languages in our \texttt{Cross-ToT} on the final performance. Following Table \ref{tab:language_distribution}, we integrate languages from low-resources to high-resources and vice versa. We also propose the same experiments with the addition of English.}
\label{fig:num_lang}
\end{figure}

English seems to lead Cross-lingual reasoning on arithmetic tasks, as shown in Section \ref{sec:en_matter}. Hence, to observe the impact of the number of languages and one specific, i.e., English, we propose two further analyses:
\paragraph{Cross-ToT in low-resources scenarios}
Integrating more languages into Cross-lingual prompting leads to better overall performance. As already observed in \cite{shi2022language,qin2023crosslingual}, increasing the number of languages improves downstream performance, as shown in Figure \ref{fig:num_lang} (average performances using the same setting proposed in Section \ref{sec:exp_set}).

As shown in \cite{malkin-etal-2022-balanced,blevins2022language}, the performances of the Large Language Models are highly correlated with the percentage of pre-training data in each language.

Following the approach proposed in \cite{qin2023crosslingual} and considering language distribution in the widely used multilingual pre-training dataset, which in our case is CommonCrawl \cite{commoncrawl2021}, we integrated languages in descending and ascending order based on their respective proportions (detailed in Table \ref{tab:language_distribution}).

Figure \ref{fig:num_lang} shows that adding more languages in high-resource contexts improves performance. However, when incorporating languages with limited resources, performance decreases as the number of languages increases (see low-resource in Table \ref{fig:num_lang}). Finally, adding English (the dominant percentage in standard corpora) to the prompting significantly enhances performance (see "+ English" lines in Table \ref{fig:num_lang}).

These findings emphasize that the number of integrated languages only partially determines the effectiveness of language integration. The amount of pre-training data for each language, especially for high-resource languages, plays a crucial role. Balancing multiple languages and considering available resources and impact is essential.

\paragraph{Cross-ToT in binary scenarios}
Moreover, we evaluate similar scenarios in low-resource settings and reproduce the same experiments using only two languages. 
In particular, we used the same setting proposed in Section \ref{sec:exp_set} by including only the target language and English in the prompt (example prompt in Appendix \ref{tab:prompt_En_target}).

From the results shown in Figure \ref{fig:En_ToT} (grey bar), using the target English-language tuple does not change the performance of high-resource languages. On the contrary, low-resource languages achieve significantly lower performance.
This second finding reinforces what was said earlier about the experiments on prompt compositions.

\subsection{Reasoning Evolution}
\label{sec:reasoning_evolution}
\begin{figure}[h]
\centering
    \includegraphics[width=0.52\textwidth]{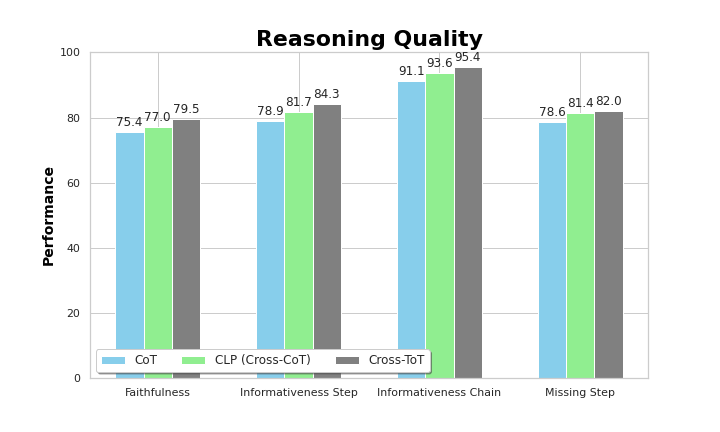}
    \caption{ The analysis of reasoning quality
between GPT-3.5 (Native-CoT) and CLP in \cite{qin2023crosslingual} and our \texttt{Cross-ToT}}
    \label{fig:reasoning}
\end{figure}

We use the framework \texttt{ROSCOE} \cite{golovneva2023roscoe} to investigate why our approach works. Hence, we evaluate the quality of the reasoning paths (implementation described in Appendix \ref{sec:reasoning_chain_APP}). As shown in Figure \ref{fig:reasoning}, our approach delivers reasoning with higher faithfulness, exhibiting better consistency with key steps during the reasoning process. Specifically, the faithfulness
score increased by 4.5 points, indicating that the model better understood the problem statement and ensured a transparent inference chain without generating irrelevant or misused information. Furthermore, we observe improvements in the Informativeness metrics for “Step” and “Chain”. It suggests that the models' reasoning, behind the alignment, could provide more well-grounded inference steps.

\begin{table}[h]
\begin{tabular}{|p{0.9\linewidth}|}
\multicolumn{1}{l}{\texttt{\textbf{XCOPA, XNLI and PAWS-X}}} \\
\hline
\texttt{Simulate the collaboration of $n$ person answering a question in their mother tongue: $L_1$ and $English$. They all start Step1 from a separate thought process, step by step, each explaining their thought process. Following Step1, each expert refines and develops their thought process by comparing themselves with others. This process continues until a definitive answer to the question is obtained. } \\
\rowcolor{gray!50}
\texttt{\uline{Basic Prompt}} \\
\hline
\end{tabular}
\caption{Our prompting approach for XCOPA, XNLI and PAWS-X. List of the \texttt{\uline{Basic Prompt}} is in Table \ref{tab:different_basic_prompts}}
\label{tab:prompt_for_other_task}
\end{table}

\subsection{The Cross-Reasoning in other tasks}
\label{sec:other_task}
Furthermore, to scale our approach, we test the applicability of \texttt{Cross-ToT} on two different task types using the same structure adapted to them as in Table \ref{tab:prompt_for_other_task}.

\paragraph{Understanding task}
We proposed our approach, Cross-ToT, on other multilingual reasoning datasets belonging to the undertandings genre. As introduced in Section \ref{sec:exp_set}, we used XNLI \cite{conneau-etal-2018-xnli} and PAWS-X \cite{yang-etal-2019-paws}. As Figure \ref{fig:performances_Understanding} shows, Cross-ToT is able to perform better in most languages. Compared to the previous SOTA obtained in CLP \cite{qin2023crosslingual}. Thus, we observed average improvements of 3.2 points on XNLI and 2.5 points on PAWS-X.

\paragraph{Commonsense Reasoning task}

We have used our approach, Cross-ToT, to an additional dataset of multilingual commonsense reasoning, as introduced in Section \ref{sec:data}. We used XCOPA as our benchmark. For comparison purposes, we considered CLP and Native-CoT proposed by \citet{qin2023crosslingual}. In Figure \ref{fig:reasoning}, we can observe that our approach has outperformed previous methods in many languages.

The results show the effective functionality of our \texttt{Cross-ToT} on different tasks. Although the method has shown appreciable increases, we continue the studies in Section \ref{sec:other_app} by observing whether adding in-context examples in the input-prompt can benefit LLMs.

\subsection{Other approaches}
\label{sec:other_app}
\texttt{Cross-ToT} can be further empowered with in-context learning. In fact, as shown in Table \ref{tab:Results_ICL}, in-context learning (ICL) techniques have achieved performant results on the downstream performance of LLMs. In particular, in further exploration of \texttt{Cross-ToT} within ICL, we conducted different experiments. 

\paragraph{From Zero- to Few-shot}
In the first experiment, we sampled 50 random instances from MGSM. Then, we replicated the experiments proposed in Section \ref{sec:exp_set}. However, we constructed the prompt by merging instances in one-shot and three-shot settings. Table \ref{tab:Results_ICL} shows that providing context makes the models more robust.

\paragraph{Performances Other Models}
\texttt{Cross-ToT} does not outperform other approaches in open-source models with fewer parameters.
Table \ref{tab:smaller} shows the performances of Llama-2-13B \cite{touvron2023llama} and Bloomz-7B \cite{muennighoff2022crosslingual}.
We hypothesize that these performances are due to the misleading behaviors observed in \cite{wei2023chainofthought} prompting CoT in models with less than 100 billion parameters. In future developments, we will continue to investigate this phenomenon.

\section{Related Work}
Large Language Models (LLMs) with billions of parameters demonstrate in-context learning and few-shot learning abilities \cite{brown2020language,wei2022emergent,min-etal-2022-rethinking} to guide LLMs to generate desired task responses, marking the advent of the prompting era and surpassing the age of the intermediate steps in algorithmic and structured reasoning \cite{roy-roth-2015-solving,ling-etal-2017-program}. 
Nevertheless, early works challenged the efficacy of few-shot techniques for empowering the prompting phase and downstream performances. In particular, \citet{yao2023tree} refined the original idea of Chain-of-Thought (CoT) \cite{wei2022emergent} by considering various reasoning paths as well known as Tree-of-Thought. 

The traditional and derivated CoT mechanisms have achieved considerable success but are limited to generating answers within a single language (i.e., English). \citet{shi2022language} proposed a multilingual evaluation that \citet{qin2023crosslingual} extended to cross-lingual scenarios. In particular, \citet{qin2023crosslingual} proposed a prompt mechanism to handle requests in any language and generate CoT specifically in English. This approach, which in our construct we called Cross-CoT has been proposed both single-phase, i.e., as a single prompt (CLP) also adopted by \cite{huang2023languages} and multi-phase (CLPS) i.e., characterized by self-consistent prompts that follow the prompting methodology proposed in \cite{qiao-etal-2023-reasoning}. Although the mechanism achieves state-of-the-art cross-linguistic reasoning steps, the single-phase promting underperforms in low-resouces languages and the multi-phase prompting characterized by a series of cascading prompts is supported far away from the zero-shot chain-of-thought concept.

In our work, we propose a method of CoT reasoning inspired. Specifically, we elicit the cross-lingual generation of a series of parallel Cross-lingual reasoning paths using a single prompt. In fact, our method is inspired by the Tree-of-Thoughts approach proposed by \cite{yao2023tree}. Hence, in a different way from previous approaches, our technique generates shared parallel reasoning paths that share the "thoughts process" delivering Self-consistent answers and reducing reasoning steps.
Our work goes beyond in the following ways:
\begin{itemize}
    \item Proposal of novel zero-shot prompting methods in cross-lingual scenarios characterized by low-resource and high-resource languages.
    \item Studying cross-lingual multi-step reasoning mechanisms using arithmetic reasoning tasks.
    \item In-depth study of the reasoning pathways provided by our prompting approach (impact of the number of languages and strongly high-resource languages).
    \item Experiments on effective functioning in commonsense reasoning and language understanding tasks.
\end{itemize}

\section{Future Works}
In future work, we intend to incorporate smaller-scale Language Models (SLMs) into our evaluations. However, the ability to produce multi-step reasoned answers is limited in SLMs. To address this, a range of techniques are emerging to align and transfer reasoning abilities between LLMs and SLMs \cite{ranaldi-freitas-2024-aligning}.

Our aim is to enhance current alignment pipelines \cite{ranaldi2023empowering,ranaldi-pucci-2023-english} to enable cross-lingual reasoning capabilities across different languages and scenarios. Including methods that emphasize the importance of language structure \cite{zanzotto-etal-2020-kermit} and uphold the foundational pillars of the NLP ecosystem \cite{app13020677}.

\section{Conclusion}

Chain-of-Thought is an outstanding prompting technique. However, the imbalance of languages in pre-training data does not always produce robust results. 
Different state-of-the-art works have proposed cross-lingual techniques to align performances obtained in different languages. They are limited to handling one language at a time or proposing multiple prompting stages, making them difficult to manage.
In this paper, we propose \texttt{Cross-ToT}, a prompting technique to elicit multi-step reasoning abilities in Cross-lingual scenarios. Hence, we elicit models to deliver answers in a Self-consistent way, collaborating to the final answer.
We have shown the functionality of our \texttt{Cross-ToT} through performance improvements obtained in a multilingual mathematical problem task.  
In addition, we have demonstrated the scalability in tasks related to commonsense reasoning and language understanding. Finally, we conducted a series of in-depth analyses in which we measured the impact brought about by low-resource vs. high-resource languages and the inclusion of English.
Our contribution aims to propose more robust models that can break down issues arising from language barriers and provide more reliable results.

\section*{Limitations}
Due to the limitations imposed by the evaluation benchmarks and the cost of the OpenAI API, we conducted tests on 16 languages in total, which only scratches the surface of the world's vast array of languages. Furthermore, our approach is based on English. It should be evaluated whether the model written in the language of the task can lead to better performance and how best to construct instructions in each language. Furthermore, we only tested the effectiveness of our method on GPT-based models (gpt-3.5-turbo). In the future, it will be worthwhile to study the generality of our model on more models, such as PaLM and Llama-2-70.

\section*{Ethics Statemets}
In our work, ethical topics were not addressed. The data used comes from open-source benchmarks, and statistics on language differences in commonly used pre-training data were obtained from official sources without touching on issues related to gender, sex, or race differences.

\bibliography{anthology,custom}

\newpage

\appendix

\clearpage

\section{Prompt}
\label{sec:prompt}
In this paper, we analyze our prompting approach, i.e., Cross-ToT, in different tasks. In Figure \ref{fig:our_proposal} we have shown the input-prompt for the MGSM \cite{cobbe2021training}. Here, we show the prompt framework for the other tasks:
\begin{table}[h]
\begin{tabular}{|p{0.9\linewidth}|}
\multicolumn{1}{l}{\texttt{\textbf{XCOPA, XNLI and PAWS-X}}} \\
\hline
\texttt{Simulate the collaboration of $n$ person answering a question in their mother tongue: $L_1$ and $English$. They all start Step1 from a separate thought process, step by step, each explaining their thought process. Following Step1, each expert refines and develops their thought process by comparing themselves with others. This process continues until a definitive answer to the question is obtained. } \\
\textbf{\texttt{\uline{Basic Prompt}}} \\
\hline
\end{tabular}
\caption{Our prompting approach for XCOPA, XNLI and PAWS-X. List of the \texttt{\uline{Basic Prompt}} is in Table \ref{tab:different_basic_prompts}}
\label{tab:prompt_for_other_task}
\end{table}

Furthermore, in Section \ref{sec:en_matter}, we proposed an experiment based on a prompt with only two languages as follows:

\begin{table}[h]
\small
\centering
\begin{tabular}{|p{0.9\linewidth}|}
\multicolumn{1}{l}{\texttt{\textbf{Binary Cross-ToT}}} \\
\hline
\texttt{Simulate the collaboration of 2 mathematicians answering a question in their mother tongue: $L_1$ and $English$. They all start Step1 from a separate thought process, step by step, each explaining their thought process. Following Step1, each expert refines and develops their thought process by comparing themselves with others. This process continues until a definitive answer to the question is obtained. } \\
\texttt{Question: [Question in Language $L_1$]} \\
\texttt{Answer: [num].} \\ 
\hline
\end{tabular}
\caption{Our prompting approach for experiment proposed in Section \ref{sec:en_matter} regarding MGSM and binary trees}
\label{tab:prompt_En_target}
\end{table}

\section{Reasoning Chain}
\label{sec:reasoning_chain_APP}
\subsection{Chain-of-Thought Quality Scoring Implementation}
The ROSCOE framework \cite{golovneva2023roscoe} incorporates multiple chain-of-thought quality metrics, with the reasoning alignment vector $\alpha$  that is  \begin{equation}
r_{align}(h \rightarrow s) = \{ \alpha_1, \alpha_2, \ldots, \alpha_N \} \in [0, 1]^N 
\end{equation}
from the $N$-step hypothesis $h = \{h_i\}_{i=1}^N$ to the source input $s$ of length $T$, where $\alpha_i$ are defined as: $r_{align}(h_i \rightarrow s) = \frac{1 + \max_{j=1}^T \cos(h_i, s_j)}{2}$

\paragraph{Faithfulness score}
The Faithfulness $(F)$ score is calculated based on
the alignment between the hypothesis steps $h$ and
the source sentences $s$. It represents the average
reasoning alignment score over the steps of reasoning:
\begin{equation}
    F = \frac{1}{N} \sum_{i=1}^{N} r_{align}(h_i \rightarrow s)
\end{equation}

The Faithfulness score serves as a measure to assess whether the model misconstrued the problem in the statement or if the reasoning chain is characterized by ambiguity, unimportance, or the misuse of information.

\paragraph{Informativness}
Informativeness-Step (Info-Step) measures the utilization of facts from the original text $s$ in the reasoning steps $h$:
\begin{equation}
    Info_{Step} = \frac{1}{2T} \sum_{t=1}^{T} r_{align}(s_t \rightarrow h) + \frac{1}{2} F
\end{equation}

Info-Step assigns a higher score to reasoning steps that strongly align with the source, showing the capacity to which the generated hypothesis includes the information from the source. Conversely, a lower Info-Step score means reasoning steps unrelated to the source sentences or overlooking the provided information in the context.

\paragraph{Informativeness Chain}
Like the Info-Step metric, the InformativenessChain (Info-Chain) metric estimates the degree of concordance between the hypothesis chain and the source. The calculation is as follows:
\begin{equation}
    Info_{Chain} = \frac{1 + \cos(h, s)}{2}
\end{equation}

\paragraph{Missing Step}
The Missing Step (Miss-Step) metric is introduced to estimate any significant lacking steps, which examines the alignment between the reference reasoning text $r = \{r_i\}^K$ and the hypothesis $h$. A miss-step is needed to meticulously assess each step in the reference and verify the existence of a similar step in the hypothesis. The metric is computed as:

\begin{equation}
  \text{Miss-Step} = \min_{i=1}^{K}(\text{r-align}(r_i, h)).
\end{equation}

\begin{table*}[t]
\section{Other Results}
\centering
\begin{tabular}{l|cccccccccc|cc}
\textbf{\# of shot- \texttt{Cross-ToT}} & \textbf{de} & \textbf{zh} & \textbf{fr} & \textbf{ru} & \textbf{sw} & \textbf{es} & \textbf{bn} & \textbf{ja} & \textbf{te} & \textbf{th} & \textbf{Avg} \\ \hline
\textbf{0-shot} & 86.5 & 84.2 & 83.9 & 83.2 & 74.3 & 84.4 & 78.7 & 79.8 & 68.7 & 74.6 & 79.8 \\
\textbf{1-shot} & 87.2 & 84.9 & 85.8 & 85.3 & 76.4 & 85.2 & 81.2 & 81.3 & \textbf{70.5} & 75.5 & 79.9 \\
\textbf{3-shot} & \textbf{88.4} & \textbf{85.7} & \textbf{87.2} & \textbf{87.5} & \textbf{77.3} & \textbf{87.3} & \textbf{82.3} & \textbf{81.5} & 70.3 & \textbf{76.9} & 83.4 \\
\hline
\end{tabular}
\caption{Accuracies (\%) on MGSM using zero-shot, one-shot and three-shot}
\label{tab:Results_ICL}

\vspace{2em}
\small
\centering
\begin{tabular}{lccccccccccc|c}
\hline
\textbf{Model} & \textbf{et} & \textbf{ht} & \textbf{id} & \textbf{it} & \textbf{qu} & \textbf{sw} & \textbf{ta} & \textbf{th} & \textbf{tr} & \textbf{vi} & \textbf{zh} & \textbf{Avg} \\
\hline
Bloomz-7B \cite{muennighoff2022crosslingual} &  &  &  &  &  &  &  &  &  &  &  &  \\
\texttt{En-CoT} & 21.8 & 24.2 & 50.6 & 41.6 & 41.4 & 48.6 & 53.8 & 38.4 & 37.6 & 47.0 & 64.2 & 42.7 \\
\texttt{CLP} \cite{qin2023crosslingual} & \textbf{49.0} & \textbf{49.6} & 58.0 & \textbf{48.8} & \textbf{50.6} & \textbf{47.6} & \textbf{57.8} & 52.0 & \textbf{50.2} & \textbf{45.2} & \textbf{54.2} & \textbf{51.2} \\
\hdashline
\texttt{\textbf{Cross-ToT}} & 48.0 & 47.3 & \textbf{58.2} & 47.8 & 49.3 & 46.4 & 55.2 & \textbf{53.1} & 50.8 & 44.2 & 50.3 & 49.5 \\
\hline
llama-2-13B \cite{touvron2023llama}  &  &  &  &  &  &  &  &  &  &  &  &  \\
\texttt{En-CoT} & 39.6 & 32.5 & 58.4 & 55.8 & 47.2 & 34.6 & 47.4 & 33.2 & 43.0 & 59.6 & 50.4 & 45.6 \\
\texttt{CLP} \cite{qin2023crosslingual} & \textbf{44.8} & 48.2 & \textbf{64.4} & \textbf{70.2} & \textbf{46.6} & \textbf{47.0} & \textbf{47.8} & \textbf{46.4} & \textbf{51.2} & \textbf{58.8} & \textbf{51.4} & \textbf{52.4} \\
\hdashline
\texttt{\textbf{Cross-ToT}} & 43.3 & \textbf{49.1} & 61.5 & 65.8 & 44.4 & 46.6 & 43.7 & 42.2 & 49.5 & 55.2 & 48.2 & 50.6 \\

\hline
\end{tabular}
\caption{Comparison of smaller open-source models on XCOPA.}
\label{tab:smaller}

\section{Prompt Table}

\centering
\begin{tabular}{lc|p{12cm}}
\hline
\hline
Benchmark & \#Test & \texttt{\uline{Basic Prompt}} \\
\hline
MGSM & 250 & Question: \texttt{\{problem\}} \\
\hline
XCOPA & 200 & Here is a premise: \texttt{\{premise\}}. What is the \texttt{\{question\}}? Help me pick the more plausible option: -choice1: \texttt{\{choice1\}}, -choice2: \texttt{\{choice2\}} \\
\hline
XNLI & 200 & \texttt{\{premise\}}. Based on the previous passage, is it true that \texttt{\{hypothesis\}}? Yes, No, or Maybe? \\
\hline
PAWS-X & 200 & Sentence 1: \texttt{\{sentence1\}} Sentence 2: \texttt{\{sentence2\}} Question: Does Sentence 1 paraphrase Sentence 2? Yes or No? \\
\hline
\hline
\end{tabular}
\caption{The basic prompt of each benchmark. \#Test denotes the number of instances in the test set that we randomly selected due to the cost constraint excepted for MGSM.}
\label{tab:different_basic_prompts}

\vspace{1cm}

\section{Number of Languages}
\begin{center}

\begin{tabular}{|l|c|}
\hline
\textbf{Language} & \textbf{Percentage} \\ \hline
English (en)      & 46.3\%              \\ \hline
Russian (ru)      & 6.0\%               \\ \hline
German (de)       & 5.4\%               \\ \hline
Chinese (zh)      & 5.3\%               \\ \hline
French (fr)       & 4.4\%               \\ \hline
Japanese (ja)     & 4.3\%               \\ \hline
Spanish (es)      & 4.2\%               \\ \hline
Other             & 23.1\%              \\ \hline
\end{tabular}
\caption{Language distribution of CommonCrawl \cite{commoncrawl2021}.}
\label{tab:language_distribution}
\end{center}

\end{table*}

\end{document}